\documentclass[%
 preprint,
 amsmath,amssymb,
]{revtex4-1}

\usepackage{graphicx}
\usepackage{amssymb}
\usepackage{setspace}
\usepackage[utf8]{inputenc}
\usepackage{array}
\usepackage{natbib}
\usepackage{amsmath}
\usepackage{multirow}
\usepackage{hyperref}% add hypertext capabilities
\usepackage{subfigure}
\usepackage{xcolor}
\usepackage{dcolumn}
\usepackage{tcolorbox}
\usepackage{bm}
\usepackage{colortbl}

\usepackage{lipsum}

\newcolumntype{a}{>{\columncolor[HTML]{DAE8FC}}c}
\newcolumntype{b}{>{\columncolor[HTML]{FFCE93}}c}

\usepackage{color}
\definecolor{redcolor}{rgb}{1.0,0.,0.}

\definecolor{bluecolor}{rgb}{0,0.,1}

\begin{document}

\preprint{}
% \title{Correlations between reference and citation features in scientific manuscripts}% Force line breaks with \\

\title{A pattern recognition approach for distinguishing between prose and poetry}

\author{Henrique F. de Arruda\footnote{This author was at the \emph{S\~ao Carlos Institute of Physics, University of S\~ao Paulo}, until 31st May 2021.},$^1$ Sandro M. Reia$^2$, Filipi N. Silva$^3$,  Diego R. Amancio$^4$ and Luciano da F. Costa$^2$}

\affiliation{
$^1$ISI Foundation, Turin, Italy \\
$^2$S\~ao Carlos Institute of Physics, University of S\~ao Paulo, S\~ao Carlos, Brazil \\
$^3$Indiana University Network Science Institute, Bloomington, Indiana 47408, USA \\
$^4$Institute of Mathematics and Computer Sciences, University of S\~ao Paulo, S\~ao Carlos, Brazil
}

\date{\today}% It is always \today, today,

\begin{abstract}
Poetry and prose are written artistic expressions that help us to appreciate the reality we live. Each of these styles has its own set of subjective properties, such as rhyme and rhythm, which are easily caught by a human reader's eye and ear. With the recent advances in artificial intelligence, the gap between humans and machines may have decreased, and today we observe algorithms mastering tasks that were once exclusively performed by humans. In this paper, we propose an automated method to distinguish between poetry and prose based solely on aural and rhythmic properties.
In other to compare prose and poetry rhythms, we represent the rhymes and phones as temporal sequences and thus we propose a procedure for extracting rhythmic features from these sequences.
The classification of the considered texts using the set of features extracted resulted in a best accuracy of $0.78$, obtained with a neural network. 
Interestingly, by using an approach based on complex networks to visualize the similarities between the different texts considered, we found that the patterns of poetry vary much more than prose.   Consequently, a much richer and complex set of rhythmic possibilities tends to be found in that modality.
\end{abstract}

\maketitle

\section{\label{sec:introduction}Introduction}
It has been frequently observed that arts and science share many characteristics, especially creativity.  Consequently, a continuous search for innovation underlies both these areas, giving rise to new approaches and conventions. At the same time, these works are typically subsumed into major areas.  While in science we have areas such mathematics, humanities, biology, etc., in arts we have styles and genres. The classification of specific works in major genres requires the respective works to share some similar characteristics.  Therefore, the organization of works of arts into genres and styles is characterized by an interesting coexistence of dissimilarity (required for innovation) and similarity (required for being grouped into a same category).  In other words, the classification of works of art needs to take into account an interplay between homogeneity (within a group) and heterogeneity (between groups).  However, even the works belonging to a same group will present some dispersion, reflecting the creativity and innovation aspects expected from works of art.  The study of these structures represent an interesting and important endeavor that has progressively been approached by using computational concepts and methods~\cite{toivonen2020computational}.

In the wide area of literature, two major areas have been typically identified: prose and poetry.  Each of these have been extensively developed along centuries, giving rise to a large number of masterpieces, while contributing substantially to human culture.   Poetry has been frequently described as a literary form emphasizing rhythm and rhymes, while prose would not involve so much attention to these two aspects.  Yet, every piece of prose will incorporate some level of rhythm and rhyme, to the point that a specific genre, namely prose poetry, has also been identified and developed.  Interestingly, while humans seem to have some natural cognitive ability to distinguish between artistic and literary styles and genres, it remains an interesting and relatively challenging question to understand in a more objective and quantitative manner the two major areas of prose and poetry.

%%%%%%%%%%%%%%%%%%%%
Some researchers have also considered the classification of poetry. For instance~\cite{jamal2012poetry}, found that Support Vector Machines can classify poems into different classes. Moreover, \cite{gopidi2019computational} found that the similarity between poetry and prose can vary according to time.
In \cite{sci2040092}, it is shown that the entropy associated to English poetry changes with time, and also that the entropy depends on the language and on the author considered.
Other researches dealt with the problem of automatically generate poetry~\cite{tikhonov2018guess,talafha2021poetry}. For instance, \cite{tikhonov2018guess} proposed long short-term memory artificial neural network with phonetic and semantic embeddings to generate stylized poetry. In the latter study, they found that the poetry generated by their method outperforms random and baseline models. Furthermore, the conversion from poetry to prose have also been studied~\cite{krishna2019poetry}.

In \cite{da2020syntonets}, the authors studied a model of the relationship musical notes in terms of harmonic series, which was represented as a complex network. In another study, researchers considered the analysis of patterns of the poetry sounds~\cite{hrushovski1980meaning}.
The differences between the creative thinking and the conceptual representations of the human mind when it comes to prose and poetry were explored in \cite{doumit2013thinking}. The authors studied poetic texts from Dylan Thomas and John Gay, and prose texts from F. Scott Fitzgerald and George Orwell. They found that poetry has a wider distribution of conceptual associations than prose, and a more complex scenario is drawn when a semantic network and a neural network are considered.

In this work, we propose a protocol for extracting the musical patterns --- namely rhymes and rhythm --- from written, artistic expressions, such as prose and poetry. First, texts are converted into temporal sequences of phones, giving us the ability to identify the existing rhymes in a given time window.
Second, we propose a set of features aimed at extracting the rhythmic patterns from the phonetic sequences, which are later used by the classification algorithms employed to discriminate between the two classes considered (prose and poetry). 
In order to avoid spurious effects on our findings, we consider prose and poetic extracts with similar sizes so as to focus on their inherent construction/structure instead of their sequence length.

Our results indicate that the classifiers we tested here were able to successfully identify the type of text under consideration with an accuracy of at least $75\%$.
Interestingly, the visualization of the similarities among the items of our corpus via complex networks reinforces the idea that the proposed algorithm for feature extraction can grasp meaningful information.
On the one hand, the networks show that prose texts are more densely connected, meaning they usually share the same rhythmic patterns. 
On the other, poetry texts are weakly connected, meaning that poetry may present a wide range of rhythmic patterns, so it is less likely to find two poetic texts with the same rhyme pattern.

The effect of the structure on the feature extraction algorithm is further explored by comparing the original poetry and prose texts with their shuffled versions.
For poetry, this experiment reveals that the shuffling of the words (while the punctuation remains fixed) results in an accuracy of about $60\%$. So, there is no evident difference between both classes in terms of the considered features. 
The accuracy value, found to be slightly higher than the null case ($50\%$), indicates that the structure alone plays a marginal role in defining poetry.
For the prose, we found that the accuracy between the classification of prose and shuffled prose is about $70\%$, meaning that in this case, the order of words is more relevant for their characterization. 

The present paper is organized as follows. 
In Section~\ref{sec:model}, we describe the materials and methods used here, including details about the dataset used in our analysis, the description of how we represent the data, the proposed method for feature extraction, the classification algorithms we use, and an overview of the network representation method for visualizing the similarities between the texts. Our findings are presented in Section~\ref{sec:results}, where we compare the texts of our corpus by looking at their basic statistics and analyze the performance of the classification algorithms. Finally, in Section~\ref{sec:conc} we offer our concluding remarks along with the perspectives for future works.

%
%\begin{figure}[ht!]
%    \centering
%    \includegraphics[width=0.70\textwidth]{imgs/schematic_v2.pdf}
%    \caption{Schematic representation of the components needed to calculate \emph{citations} and \emph{references} diversity.}
%    \label{fig:schematic}
%\end{figure}

\section{Materials and methods} \label{sec:model} 

\begin{figure*}[!ht]
  \centering
    \includegraphics[width=\textwidth]{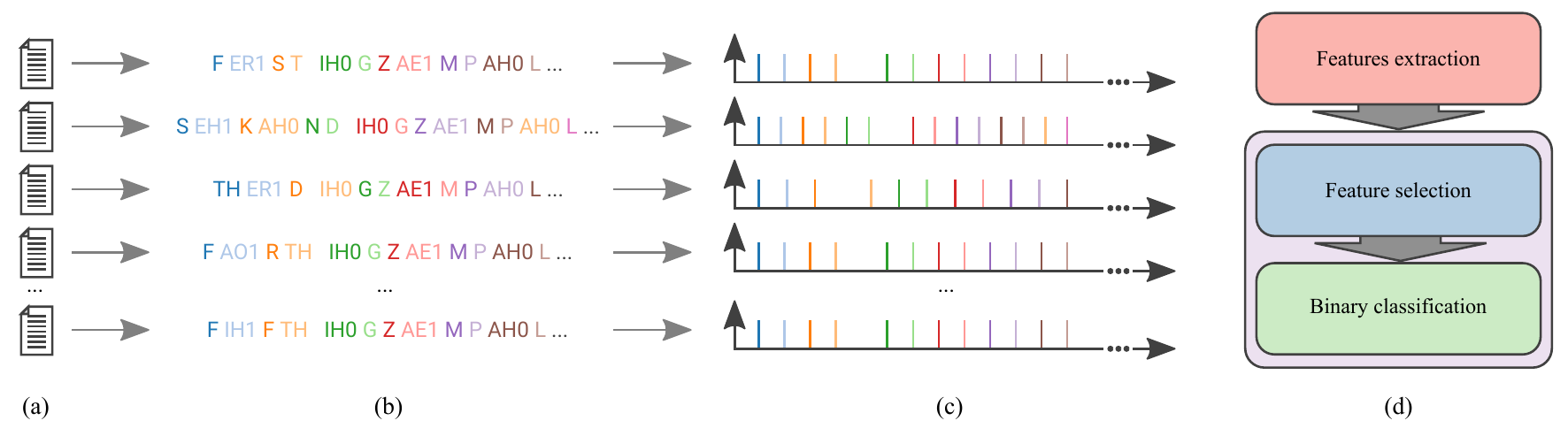}
  \caption{Pipeline of the proposed analysis. (a) Set of texts to be analyzed. (b) The texts are converted into sequences of phones. (c) The phone sequences are represented as sequences of phone repetitions. (d) The sequences obtained in (c) are classified into poetry and prose.}
  \label{fig:pipeline}
\end{figure*}

In this section, we present the employed datasets and the methodology used to represent texts as sequences. Fig.~\ref{fig:pipeline} illustrates the proposed pipeline of analysis. The employed texts, as well as the used dictionary of phones are described in Section~\ref{sec:dataset} (see Fig.~\ref{fig:pipeline}(a)~and~(b)). In Section~\ref{sec:data_representation}, we describe the methodology for representing the data and the measured features, as illustrated in Fig~\ref{fig:pipeline}(c). Furthermore, in Section~\ref{sec:classifiers}, we describe the feature selection and the classifiers used to classify sequences into prose or poetry. Afterward, each text is characterized by a set of features extracted from the corresponding sequence of phones. Since some features might not be relevant to discriminate between poetry from prose writing styles, we used a feature selection algorithm to identify the ones that contribute the most to our goal.
These methodological steps are described in Section~\ref{sec:classifiers} and illustrated in Fig. \ref{fig:pipeline}(d).

\subsection{Employed data} \label{sec:dataset}

The dataset considered here comprises extracts of texts with similar length. The majority of the samples are obtained from the Project Gutenberg website \cite{Gutenberg}, which is an online library that makes available over $60,000$ ebooks. In particular, the \emph{poetry} corpus is equally composed of odes, ballads and sonnets from different books, in a total of $60$ samples. The \textit{prose} corpus encompasses the same number of samples of technical, novel books, and pieces of news ($20$ of each). However, the latter were obtained from the Brown Digital Repository \cite{Brown}. More specifically, we select pieces of texts from the $20$ first texts in the category of \emph{news}. More details regarding the dataset characteristics are shown in Section~\ref{sec:resut_basic}.

Words from the text samples (Fig. \ref{fig:pipeline}(a)) are tokenized with the natural language toolkit (NLTK) python package \cite{bird2009natural}, and the tokens are converted into phones with the use of the pronouncing python library \footnote{https://github.com/aparrish/pronouncingpy} (Fig. \ref{fig:pipeline}(b)).
The pronouncing library is based on the \emph{Carnegie Mellon University Pronouncing Dictionary}~\footnote{http://www.speech.cs.cmu.edu/cgi-bin/cmudict} that is an open-source pronunciation dictionary for North American English containing about 134 thousand words and their pronunciations. It is useful for speech recognition since it maps words to their pronunciations in the ARPAbet phoneme set \cite{mines1978frequency,oh2005ensemble,zegers1998speech}, having 39 phones for standard English pronunciation.

As a result, each text is represented as a sequence of phones as shown in Fig.~\ref{fig:pipeline}(c), in which each colored vertical bar represents a different phone. The intervals between phones take into account the unities of time presented in Table \ref{tab:definitions}, which is further explained in the next section.

\subsection{Data representation} \label{sec:data_representation}

\begin{figure*}[!ht]
  \centering
  \includegraphics[width=1\textwidth]{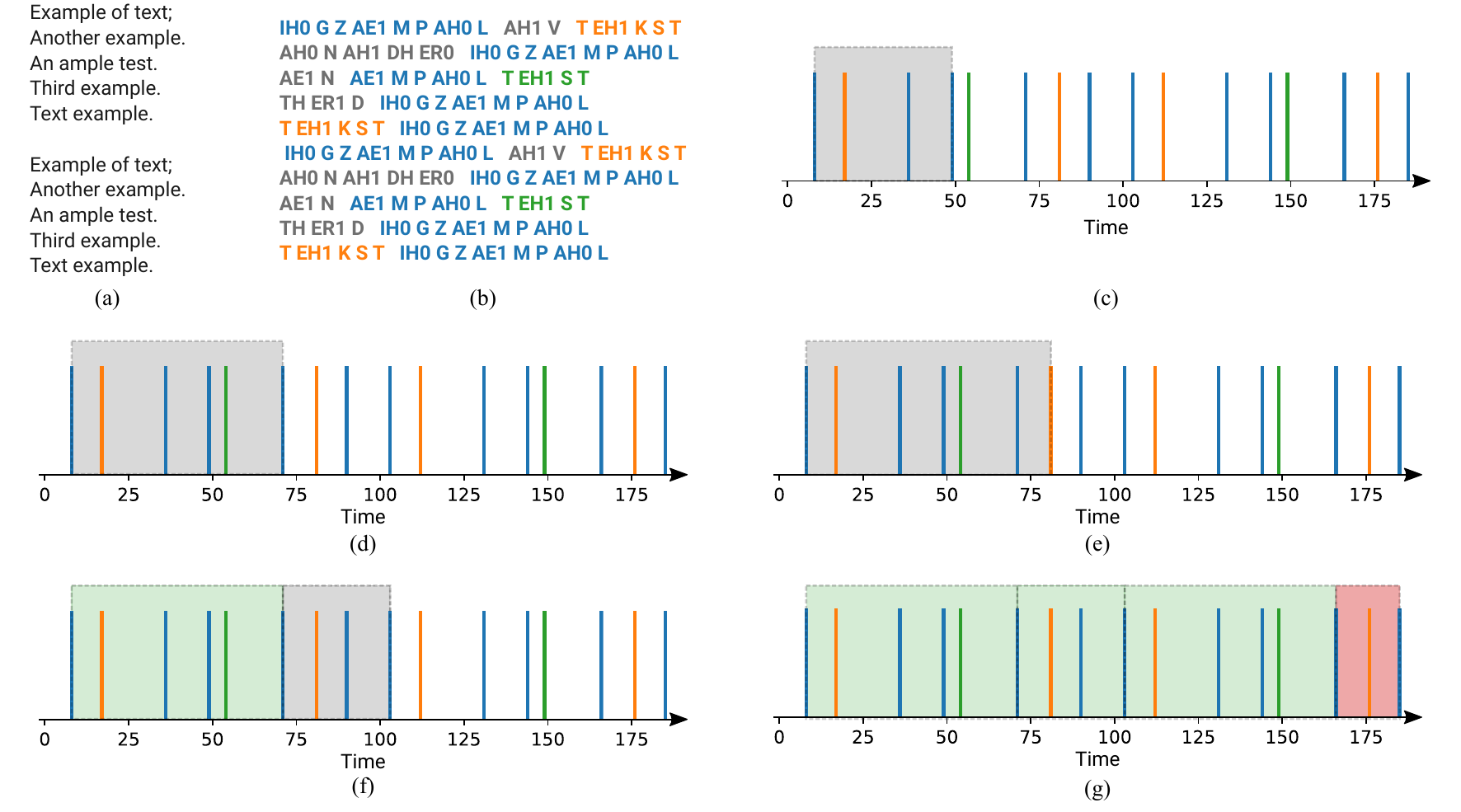}
  \caption{Example of the proposed approach. (a) Text to be processed. (b) Tokens converted into phones and colors representing rhymes. The gray phones, non-related to rhymes, are not considered in the method. Next, from (c) to (g), an example of how the windows are defined is shown, in which the green windows are detected. The red window, shown in (g), is not considered. Here we employ the following parameters: $L_0=2$ and $\Delta=0.2$.}
  \label{fig:methodology}
\end{figure*}

A music representation was proposed in~\cite{sound2021costa} capable of quantifying rhythmic and aural patterns. We propose a similar methodology, but here we aim at comparing texts by considering their rhyme and rhythmic structure. For this purpose, we introduce a methodology based on \emph{phones} and \emph{rhymes}. The central concept here is to characterize the temporal distribution of rhymes, which we believe can be related to rhythm in sounds. Fig.~\ref{fig:methodology} illustrates our approach. 

The representation is created for each text separately. The entire algorithm is shown in Fig.~\ref{fig:code}. The method starts with a text (one example of text is shown in Fig.~\ref{fig:methodology}(a)). First, a simple pre-processing step is executed, in which consecutive break lines are reduced into a single break line.
Next, we identify all tokens in the text, which include punctuation marks, numbers, and line breaks. For all tokens, the respective phones are found; see Fig.~\ref{fig:methodology}(b). No phones are attributed for punctuation and other tokens without a respective phone in the dictionary. The set of tokens preceded by fixed punctuation are selected, and all words that rhyme with these tokens identified, as represented in Fig.~\ref{fig:methodology}(b). This set of words is henceforth referred to as \emph{rhyme words}. In this study, we considered the punctuation set as: ``.'', ``:'', ``;'', ``!'', and ``?''. This punctuation set is called \emph{rhythm punctuation}. Further details regarding the choice of \emph{rhythm punctuation} are shown in Section~\ref{sec:resut_basic}.

\begin{figure}
    \centering
    \begin{tcolorbox}
    
    \begin{flushleft}
    \textbf{Data:} 
    
    $T$; \textcolor{gray}{//input text.}
    
    $L_0$; \textcolor{gray}{//Initial number of pair of consecutive signals.}
    
    $\Delta$; \textcolor{gray}{//parameter $\Delta$.}
    
    \textbf{Begin}
    \begin{itemize}
    	\item[] $W$; \textcolor{gray}{//Starts as an empt list.}
        \item[] $T$ $\gets$ \emph{preprocess}($T$);
        \item[] \emph{tokens} $\gets$ \emph{tokenize}($T$); \textcolor{gray}{// tokenize the text.}
        \item[] \emph{phones} $\gets$ \emph{find\_phones}(\emph{tokens})
        \item[] $t_s$ $\gets$ find\_time\_series(\emph{tokens}, \emph{phones}); 
        \item[] $w$ $\gets$ find\_initial\_window($t_s$, $L_0$); \textcolor{gray}{//$w$ stores the initial and final positions of the window.}
        \item[] $T_w$ $\gets$ find\_time\_differences($w$, $t_s$);
        \item[] \textbf{while} there are possible windows \textbf{do}:
        \begin{itemize}
            \item[] $w_2$ $\gets$ find\_the\_next\_window($t_s$, $w$);
            \item[] $T_{w_2}$ $\gets$ find\_time\_differences($w_2$, $t_s$);
            \item[] \textbf{if} $|\text{\emph{cv}}(T_{w}) - \text{\emph{cv}}(T_{w_2})| > \Delta$ \textbf{then}:
            \begin{itemize}
                 \item[] $w$ is inserted in $W$;
                 \item[] $w$ $\gets$ start\_new\_window($t_s$, $L_0$, $w$); \textcolor{gray}{//the final position of the previous window ($w$) is the first in the new window.}
        		\item[] $T_w$ $\gets$ find\_time\_differences($w$, $t_s$);
            \end{itemize}
             \item[] \textbf{else}: 
             \begin{itemize}
                \item[] $w$ $\gets$ $w_2$;
                \item[] $T_w$ $\gets$ $T_{w_2}$;
            \end{itemize}
         \end{itemize}
    \end{itemize}
    \textbf{End}
    \end{flushleft}
    \end{tcolorbox}
    
    \caption{Pseudocode of the proposed algorithm for clustering the signals.}
    \label{fig:code}
\end{figure}

We define the time scale in terms of the number of phones for each token, so that one phone correspond to a single time unit. In the case of words that are not part of the dictionary of phones, we added one unit of time. Furthermore, one unit of time is added between consecutive words. For punctuation and break lines, we considered the dictionary shown in Table~\ref{tab:definitions}.

\begin{table}[!ht]
    \centering
    \begin{tabular}{cc}
        \hline
        \textbf{Symbol} & \textbf{Unities of time} \\
        \hline
        , & 3\\
        . & 4\\
        ; & 4\\
        ! & 5\\
        ? & 5\\
        - & 5\\
        -- & 5\\
        \emph{break line} & 1\\
        \hline
    \end{tabular}
    \caption{Unities of time defined for the considered \emph{rhythm punctuation}.}
    \label{tab:definitions}
\end{table}

By considering the position of the phones in the time scale, a signal is assigned in the position of the last phone of each \emph{rhyme word}. The rhymes are discriminated by type; all the words that rhyme are represented by the same signal type. The step of the definition of the time series, which includes the information of rhymes and the time unities, is represented by the function \emph{find\_time\_series} of Fig.~\ref{fig:code}. One example of temporal representation is shown in Fig.~\ref{fig:methodology}(c), and the rhymes are discriminated by considering different colors. 

With the rhyme sequence in hand, we clustered the signals with a similar variety of gaps into windows. The clustering defined by the while loop of the algorithm is shown in Fig.~\ref{fig:code}. This method begins with a window that incorporates $L_0$ pairs of consecutive signals with the same type. Note that it is possible to have signal pairs with different rhyme types. The window begins at the first signal. In the example of Fig.~\ref{fig:methodology}(c), we used $L_0=2$, and the two pairs are defined between blue signals. By considering this window, the coefficient of variation is calculated for the time differences between signals, which is defined as $ \text{\emph{cv}}(T_w) = {\sigma}/{\mu},$
%\begin{equation}
% %\end{equation}
where $T_w$ is a vector with the considered time differences of the window $w$, and $\mu$ and $\sigma$ are average and standard deviation of the $T_w$, respectively. In Fig.~\ref{fig:code}, the coefficient of variation is represented by the function \emph{cv}.

For each step, new signals are incorporated into the window until another pair of related signals is obtained (given by the function \emph{find\_next\_window} in the algorithm of Fig.~\ref{fig:code}). This new window $w_2$ gives rise to another set of time differences $T_{w_2}$. Next, $cv$ is calculated for both $T_{w}$ and $T_{w_2}$. Another variable, $\Delta$, is defined to represent the maximum difference between $T_{w}$ and $T_{w_2}$. More specifically, if  $|\text{\emph{cv}}(T_{w}) - \text{\emph{cv}}(T_{w_2})| > \Delta$ is reached, the process stops, the window $w$ is stored in $W$, and a new window starts from the last signal of $w$. Otherwise, $T_{w}$ is replaced by $T_{w_2}$ and the process resumes into its signal-pairing stage. The algorithm finishes when there is no possibility to create a new window. Furthermore, if a window does not finish at the end of the algorithm, it is not added to $W$.

In the example of Fig.~\ref{fig:methodology}, the first window begins with two pairs of the blue signal, as shown in Fig.~\ref{fig:methodology}(c). The next possible signal is tested in Fig.~\ref{fig:methodology}(d), and the difference of \emph{cv} is lower than $\Delta$. So, the next possible window is tested (see Fig.~\ref{fig:methodology}(e)). In this case, the orange signal is added, which gives rise to a relatively high time difference. Consequently, the window, $w$, finishes, and this signal is not added to $W$. Fig.~\ref{fig:methodology}(f) shows the first defined window in green, and the start of a new window, in gray. The signal taking part of the end of a window is the first in the next one. This process is repeated for all possible signals, and three windows in green are created (see Fig.~\ref{fig:methodology}(g)). In red, we illustrate the signals that did not give rise to a new window. 

\subsection{Data characterization}
We propose some metrics to analyze the time sequences and the window sizes identified in the previous section.
These measurements are employed to characterize the texts and, as the next step, to compare between \emph{poetry} and \emph{prose}. In the following, we itemize the employed measurements:

\begin{itemize}
    \item $\mu_l$: mean of the time intervals between the first and the latter signal in the windows;
    \item \emph{cv}($l$): coefficient of variation of the time intervals between the first and the latter signal in the windows;
    \item $\mu_d$: mean of the differences between pairs of consecutive signals of the same type, which is calculated for each window;
    \item $\sigma_d$: the standard deviation of the differences between pairs of consecutive signals of the same type, which is calculated for each window;
    \item $\mu_l \times {cv}(l)$: in order to understand if there is a relationship by considering both quantities together, we also considered $\mu_l \times {cv}(l)$.
\end{itemize}

Because the measurements are computed for each detected window, we considered both the average and standard deviation of the described features to characterize documents. 

\subsection{Classification} \label{sec:classifiers}
In order to identify the characteristics associated with each type of text, we used feature selection algorithms. Furthermore, the quality of this set of features is quantified by considering some different classifiers. All these methodologies are described in this section.

As an attribute selection, we use the \emph{Information gain}~\cite{azhagusundari2013feature,kraskov2004estimating}, which is based on information theory. This is supervised approach and consists in a comparison between the employed feature with the classes. In order to quantify the relationship between features and classes, the normalized mutual information (NMI)~\cite{kraskov2004estimating} is calculated. All the features are ranked according to their NMI. We chose this approach since the features are computed separately. In this fashion, we can better understand their relationship with the obtained rhyme sequences.

We use five classifiers based on different assumptions. Thus, we can identify if the obtained results are consistent among different classification techniques~\cite{amancio2014systematic}. In the following, we list the employed classifiers, along with the considered parameters:
\begin{itemize}

    \item \emph{LDA}: the Linear Discriminant Analysis (e.g.~\cite{friedman2001elements}) attempts to find a linear combination of features that can be used to classify two or more classes. In this case, we considered a single LDA dimension;
    
    \item \emph{RF}: the Random Forest method (e.g.~\cite{breiman2001random}) considers an ensemble of decision trees that are merged to yield a more accurate prediction or classification. We set the maximum depth of the tree as $2$ and the random state as $0$. The remaining parameters were set as default;
    
    \item \emph{$K$NN}: the $K$ Nearest Neighbors classifier (e.g.~\cite{bentley1975multidimensional}) basically assumes that similar objects are closer to each other according to some metrics (such as Euclidean distance in multidimensional space). A parameter $K$ consists of the number of considered neighbors. Here we set $K=5$, and the remaining parameters as default;
    
    \item \emph{SVM}: the Support Vector Machine (e.g.~\cite{wu2004probability}) tries to find the right hyper-planes that maximizes the distance between it and the objects in the training set. We employed a linear kernel, and other parameters are set as default;
    
    \item \emph{MLP}: the Multi-layer Perceptron (e.g.~\cite{hinton1990connectionist}) is a multilayer artificial neural network. We set the maximum number of iterations and the hidden layer sizes as 10,000 and 40, respectively. The remaining parameters were set as default.

\end{itemize}
Because the classifiers we used here are fundamentally different from each other, we believe the individual results we obtained are complementary thus leading to a better perspective of the classification problem. All the features were standardized before the classification.

In order to avoid overfitting we use the leave-one-out as cross-validation~\cite{kohavi1995study}. More specifically, the training set is defined with all samples, except one considered the test. The same process is repeated, and, in separated steps, all features are considered as being the training set. We consider the standardization, attribute selection, and classification model fitted only with the training set in this process. All the methods presented in this section were implemented by using the scikit-learn~\cite{scikit-learn} in Python language.

\subsection{Networked approach}
\label{sec:network}

In order to better understand the relationship between the analyzed classes, we compare the proposed representation by using a network-based approach. More specifically, we visualize the similarity texts by mapping the corpus into a complex network~\cite{comin2020complex}.
%by considering the similarity between all pairs of nodes. 
In this case, each node is a text, and the edges are weighted according to the similarity between them. We considered the feature vectors obtained from each text, and the similarities are calculated for all pairs of vectors as the cosine similarity~\cite{silva2016using,gupta2014introduction}. The obtained network is visualized by using a \emph{force directed algorithm}~\cite{fruchterman1991graph}, implemented by~\cite{silva2016using}. In order to better understand the relationship among samples, we remove the edges with weights lower than a threshold $\tau$.

\section{Results and discussions} \label{sec:results}

In this section, we present the results regarding the similarities between poetry and prose. We considered the performance of the classification algorithms to discriminate \emph{poetry} from \emph{prose} based on the set of proposed rhythmic features. For this purpose, we begin analyzing the dataset and describing a few basic statistics in Section~\ref{sec:resut_basic}.

Once the poetry and prose corpus are characterized, we address the relationship between them in Section~\ref{sec:conclusionB}, in which we show that the features collection/extraction method we are proposing can capture rhythmic patterns since the Precision, Recall, and Accuracy are higher than the random baseline ($0.50$). Finally, in Section~\ref{sec:result_random_texts}, we propose a null model to explore the robustness of our findings with respect to the text structure and the words chosen by the authors.

\subsection{Dataset analysis and basic statistics}
\label{sec:resut_basic}

\begin{figure}[!ht]
  \centering
    \subfigure[]{\includegraphics[width=0.38\textwidth]{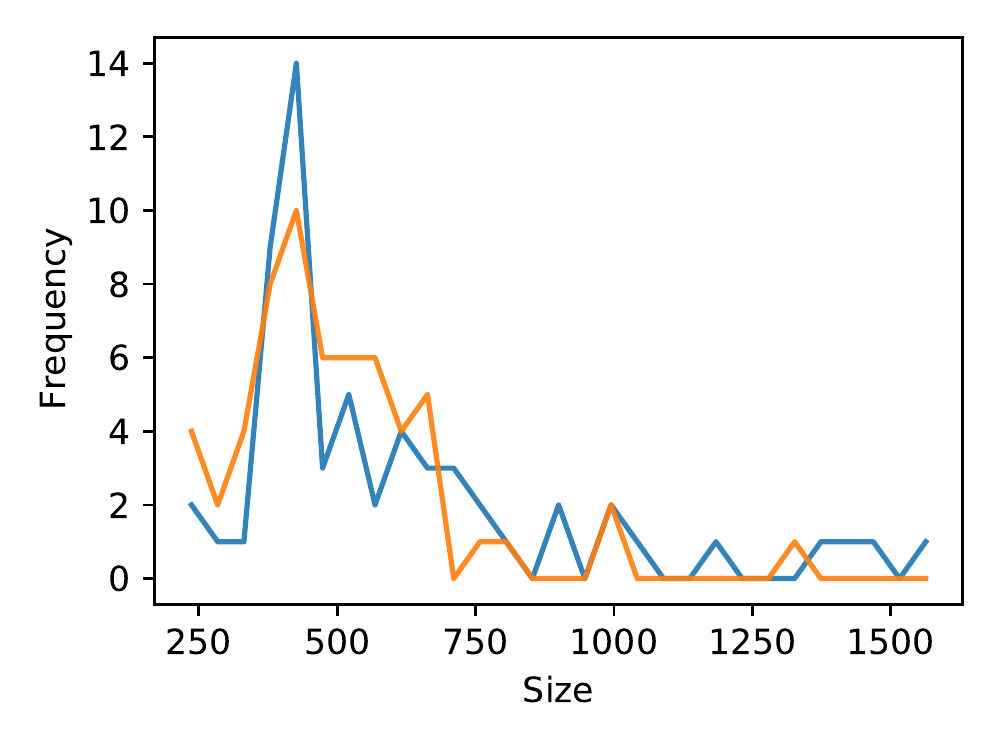}}
    \subfigure[]{\includegraphics[width=0.38\textwidth]{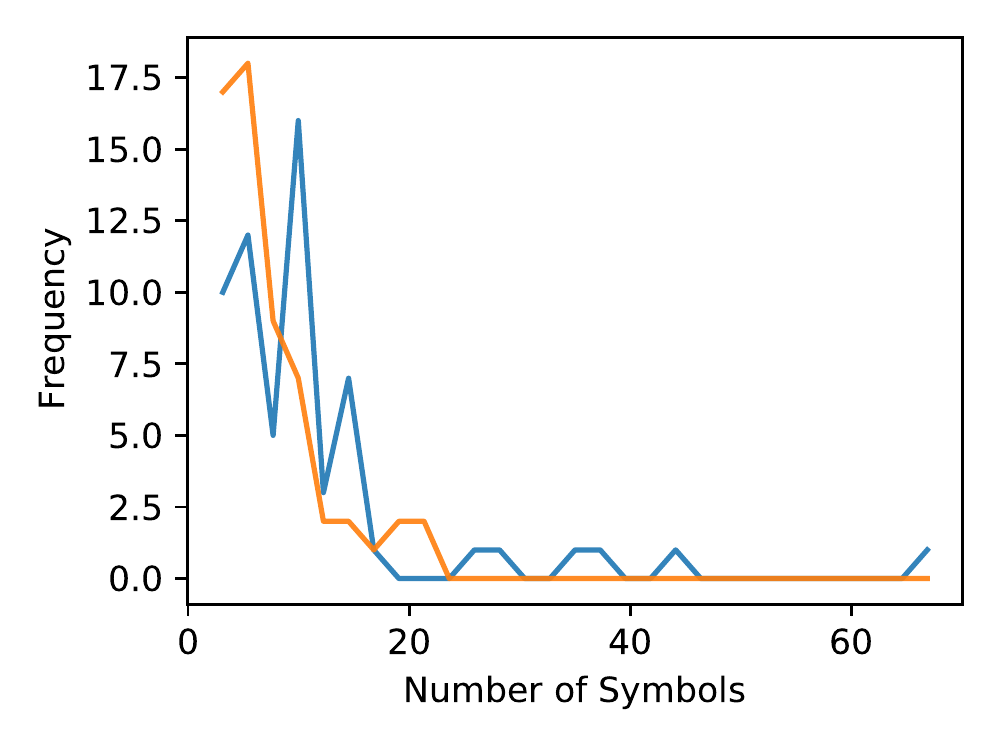}}
    \subfigure[]{\includegraphics[width=0.38\textwidth]{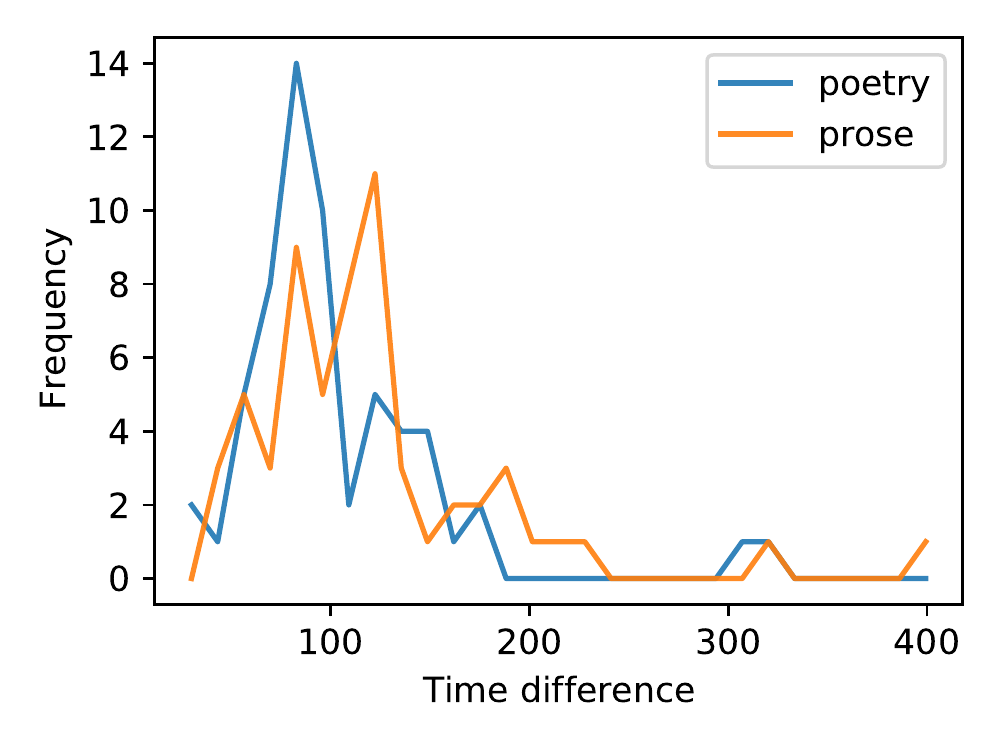}}
    
  \caption{Histograms of the basic statistics. (a) Sizes of all proses and poetries in terms of the number of phones and characters. (b) Histogram of \emph{rhythm punctuation}. (c) Histogram of the average differences between the consecutive considered punctuation set (``.'', ``:'', ``;'', ``!'', and ``?''). In this case, we considered the difference in terms of the number of characters.}
  \label{fig:statistics}
\end{figure}

Before performing the analysis regarding the data representation (described in Section~\ref{sec:data_representation}), we briefly describe a few basic information regarding the employed data. First, in order to certify that the text length is not influencing the analysis, we select texts for both classes with similar numbers of phones, as shown in Fig.~\ref{fig:statistics}(a). Another essential piece of information is the set of punctuation to be considered as the \emph{rhythm punctuation}. In this case, we searched for a particular subset of punctuation symbols (from the set shown on Table~\ref{tab:definitions}) whose frequency is similar for prose and poetry. Fig.~\ref{fig:statistics}(b) illustrates the histogram by considering the following set of symbols: ``.'', ``:'', ``;'', ``!'', and ``?''. 

Similar results were found for the comparison between the interval between the considered punctuation. In order to consider the simplest statistics possible, we employ the differences in terms of number of characters.  As can be seen in Fig.~\ref{fig:statistics}(c), the distributions were found to be similar for both classes.

All in all, the results shown in Fig.~\ref{fig:statistics} shows that the distributions for both classes are similar. So, by considering the frequencies and punctuation presented here, the employed dataset seem not to influence the results shown in the comparisons described in the following sections.

\begin{figure}[!ht]
  \centering
    \subfigure[]{\includegraphics[width=0.38\textwidth]{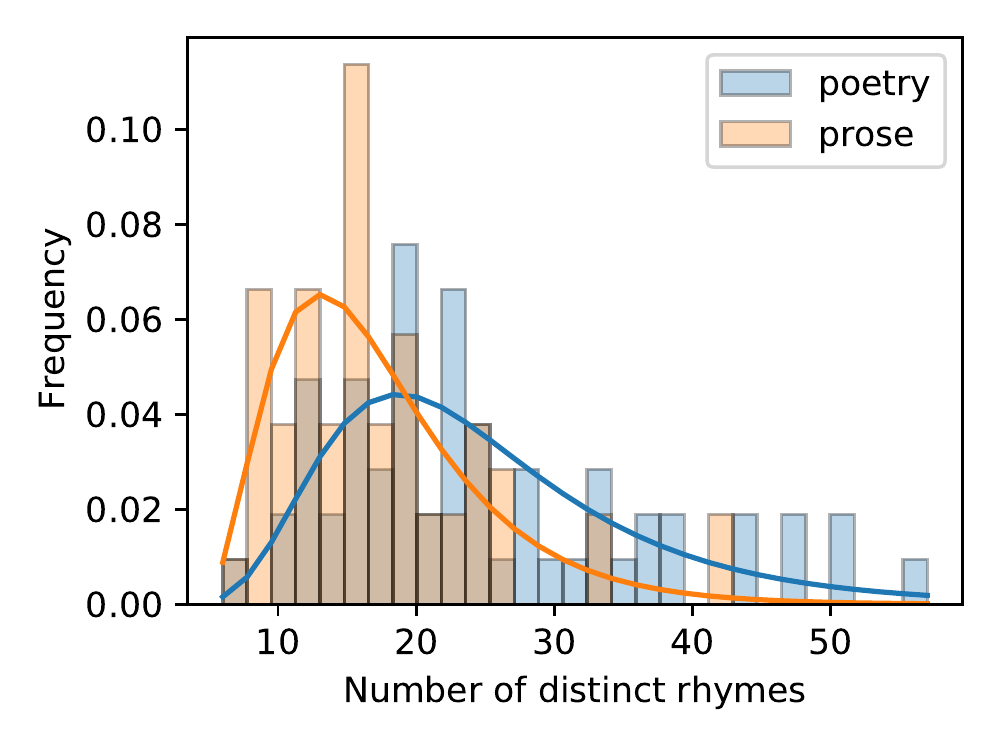}}
    \subfigure[]{\includegraphics[width=0.38\textwidth]{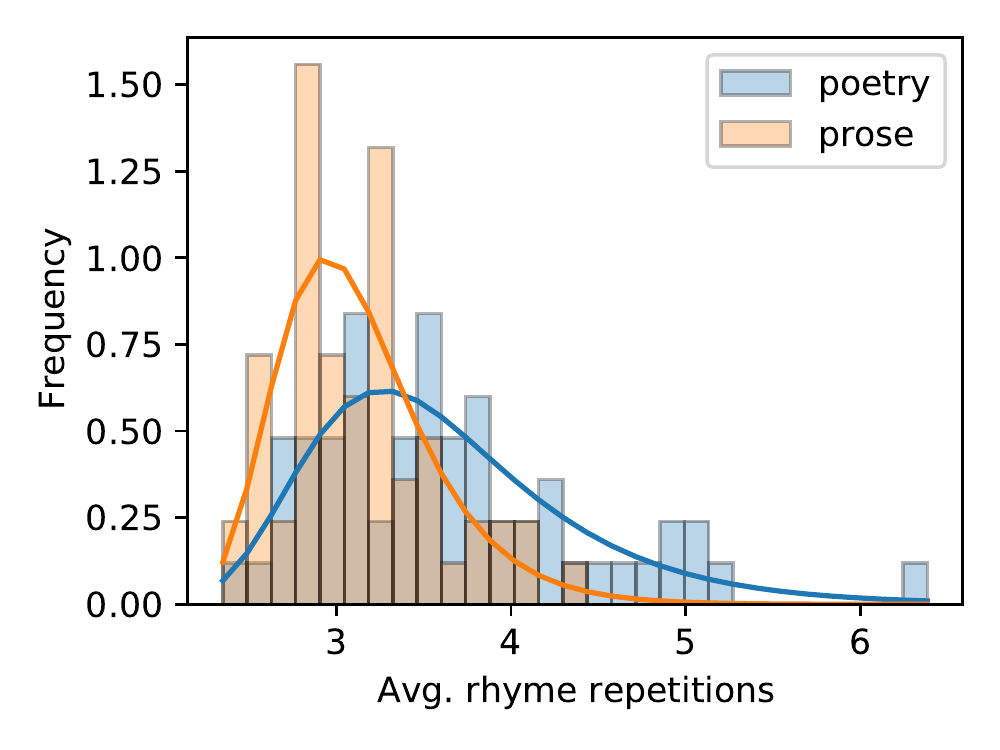}}
    
  \caption{Histograms representing patterns of rhymes. The lines represent fitting with Weibull distributions.}
  \label{fig:rhyme_statistics}
\end{figure}

Since our methodology is based on rhymes, we compared both classes in terms of their distributions (see Fig.~\ref{fig:rhyme_statistics}). More specifically, in Fig.~\ref{fig:rhyme_statistics}(a), we plot the histograms of the number of distinct rhymes, and in Fig.~\ref{fig:rhyme_statistics}(a) the average number of rhyme repetitions. In contrast with the previously presented results, here, the histograms are visually different. To show these differences, for each histogram, we fit a curve of the Weibull distribution~\cite{rinne2008weibull}. For both measurements, it is possible to note that the histograms regarding prose tend to be more concentrated on the left side of the plot. Furthermore, poetry tends to have a more spread pattern of rhymes.

\subsection{Comparison between poetry and prose}
\label{sec:conclusionB}
For each text, we create the respective phone time sequences as described in Section~\ref{sec:data_representation}. In order to better understand the differences between the classes, we classified the sequences with the methods presented in Section~\ref{sec:classifiers}. We start by using the \emph{information gain} as an attribute selection to rank the features according to their relevance. 
For all tests we considered the following set of parameters: $L_0 \in \{2, 5, 10, 15, 20\}$ and $\Delta \in \{0.01, 0.05, 0.10, 0.15, 0.20\}$.

\begin{figure}[t!]
  \centering
    \includegraphics[width=0.49\textwidth]{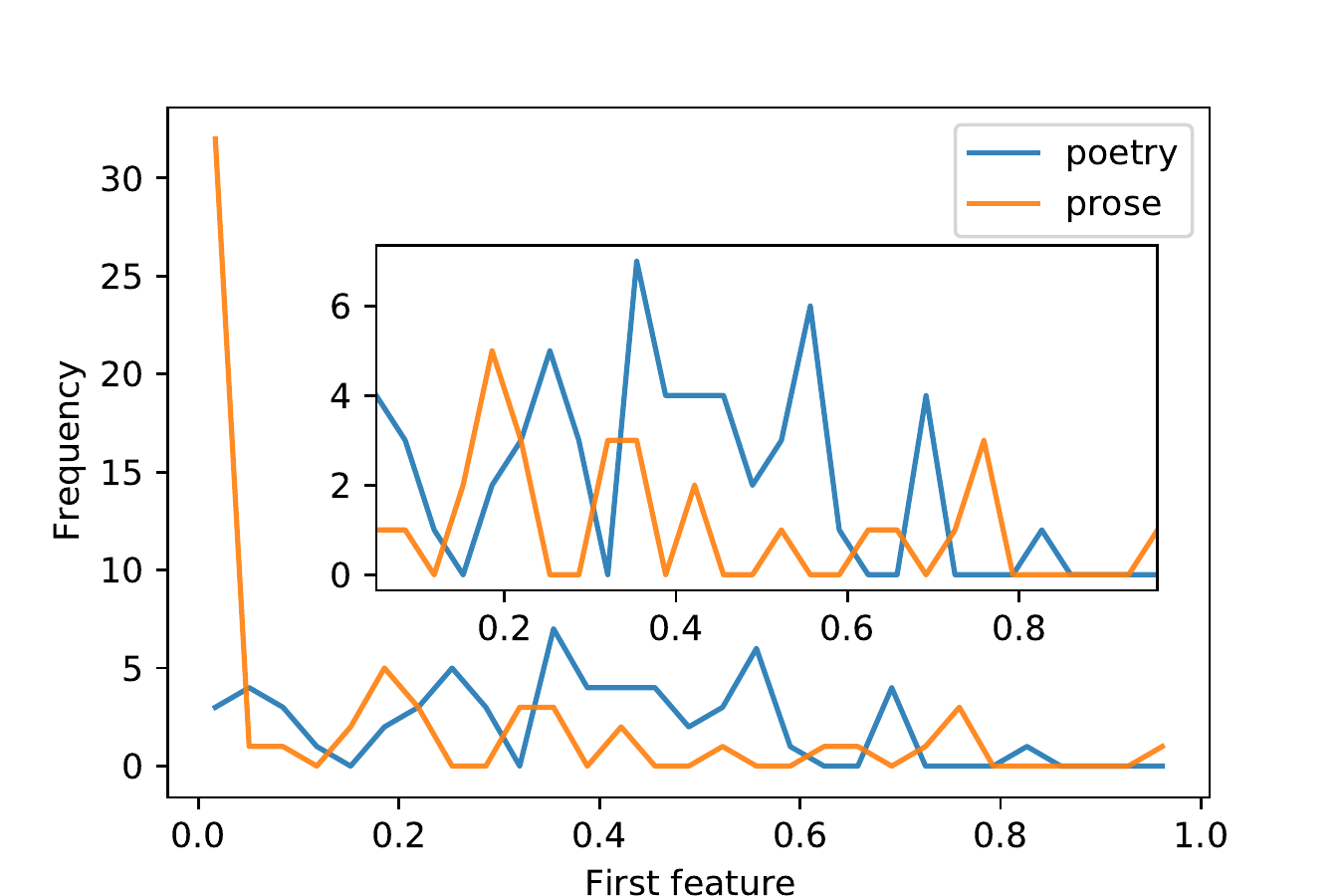}
  \caption{Histogram of the average of \emph{cv}$(l)$, which was considered by \emph{information gain} as being the most important. The employed parameters are: $L_0=2$ and $\Delta=0.1$. The inset depicts a zoom for values higher than zero.}
  \label{fig:feature0}
\end{figure}

In Fig.~\ref{fig:feature0} we show the frequency distribution of the most relevant feature across the poetry and prose corpus.
While the average $cv(l)$ is spread in a wide range of values in the poetry corpus, this metric is highly concentrated close to zero for the prose texts. 
It means that the proposed algorithm for clustering signals, described in Section~\ref{sec:data_representation}, was not able to detect a window due to one of the following reasons: (i) a pair of signals of the same type were not found in the whole text or (ii) the time distances are too regular and, consequently, the method finished without enclosing a window. Similar results were found for the other features with high values \emph{information gain}.

In order to compare the classes, we employ all the classification methods proposed in Section~\ref{sec:classifiers}. Because we are interested in the information obtained by the proposed features, we test the classifiers to obtain the best accuracy while using the smallest number of features. More specifically, we executed the classifiers for different numbers of features, from $3$ to $50$ features, which were ordered according to the \emph{information gain} attribute selection.

Table~\ref{tab:poetry_prose} illustrates the results obtained for the classifications. Interestingly, the obtained values of accuracy are similar. In spite of its simplicity, LDA obtained a relatively high accuracy using only the $n_f = 3$ most relevant features, being outperformed only by the MLP with $n_f = 14$.
However, the variation of the results of precision and recall were found to be higher than for the values of the accuracy. In the case of LDA, we observe the highest value of precision in the classification of poetry. In this case, a high value of precision means that when a text is classified as poetry, there is a high chance that the classification is correct. 
On the other hand, a high value of recall means that the classification method is correctly classifying all the poetry texts, which is the case of SVM.
In the case of prose, SVM presents the highest precision, and LDA presents the highest recall.

% --------------------------------------------------------------%
% ------------- Comparacao entre poetry and prose ------------- %
% --------------------------------------------------------------%
\begin{table}[t!]
%\resizebox{0.9\textwidth}{!}
{%
\begin{tabular}{ccccccc}
\hline
                             &                        & \multicolumn{2}{c}{Precision}                                                 & \multicolumn{2}{c}{Recall}                                                    &                            \\
\multirow{-2}{*}{Classifier} & \multirow{-2}{*}{$n_f$} & poetry                                & prose                                 & poetry                                & prose                                 & \multirow{-2}{*}{Accuracy} \\ \hline
LDA                          & \textbf{3}                      & \cellcolor[HTML]{DAE8FC}\textbf{0.81} & \cellcolor[HTML]{FFCE93}0.74          & \cellcolor[HTML]{DAE8FC}0.70          & \cellcolor[HTML]{FFCE93}\textbf{0.83} & 0.77                       \\
RF                           & 35            & \cellcolor[HTML]{DAE8FC}0.73          & \cellcolor[HTML]{FFCE93}0.77          & \cellcolor[HTML]{DAE8FC}0.78          & \cellcolor[HTML]{FFCE93}0.72          & 0.75                       \\
KNN                          & 4                      & \cellcolor[HTML]{DAE8FC}0.73          & \cellcolor[HTML]{FFCE93}0.78          & \cellcolor[HTML]{DAE8FC}0.80          & \cellcolor[HTML]{FFCE93}0.70          & 0.75                       \\
SVM                          & 13                     & \cellcolor[HTML]{DAE8FC}0.72          & \cellcolor[HTML]{FFCE93}\textbf{0.80} & \cellcolor[HTML]{DAE8FC}\textbf{0.83} & \cellcolor[HTML]{FFCE93}0.68          & 0.76                       \\
MLP                          & 14                     & \cellcolor[HTML]{DAE8FC}0.79          & \cellcolor[HTML]{FFCE93}0.76          & \cellcolor[HTML]{DAE8FC}0.75          & \cellcolor[HTML]{FFCE93}0.80          & \textbf{0.78}              \\ \hline
\end{tabular}%
}
\caption{Performance of classifiers LDA, RF, KNN, SVM and MLP on poetry and prose extracts.}
\label{tab:poetry_prose}
\end{table}
% --------------------------------------------------------------%
% --------------------------------------------------------------%
% --------------------------------------------------------------%

Despite the differences in performance, the rhythmic-based features were found to properly describe the two types of text. More specifically, independently of the nature of the employed classifier, relatively high values of accuracy were obtained. It is important to highlight that the aim of this paper is not to propose features that outperforms competing approaches in classification. Here we are more interested in demonstrating that the characteristics of rhythm can be measured. Departing from the premise that poetry and prose have different rhythms when read, our method successfully captured these differences. 

\begin{figure}[!ht]
  \centering
    \subfigure[]{\includegraphics[width=0.45\textwidth]{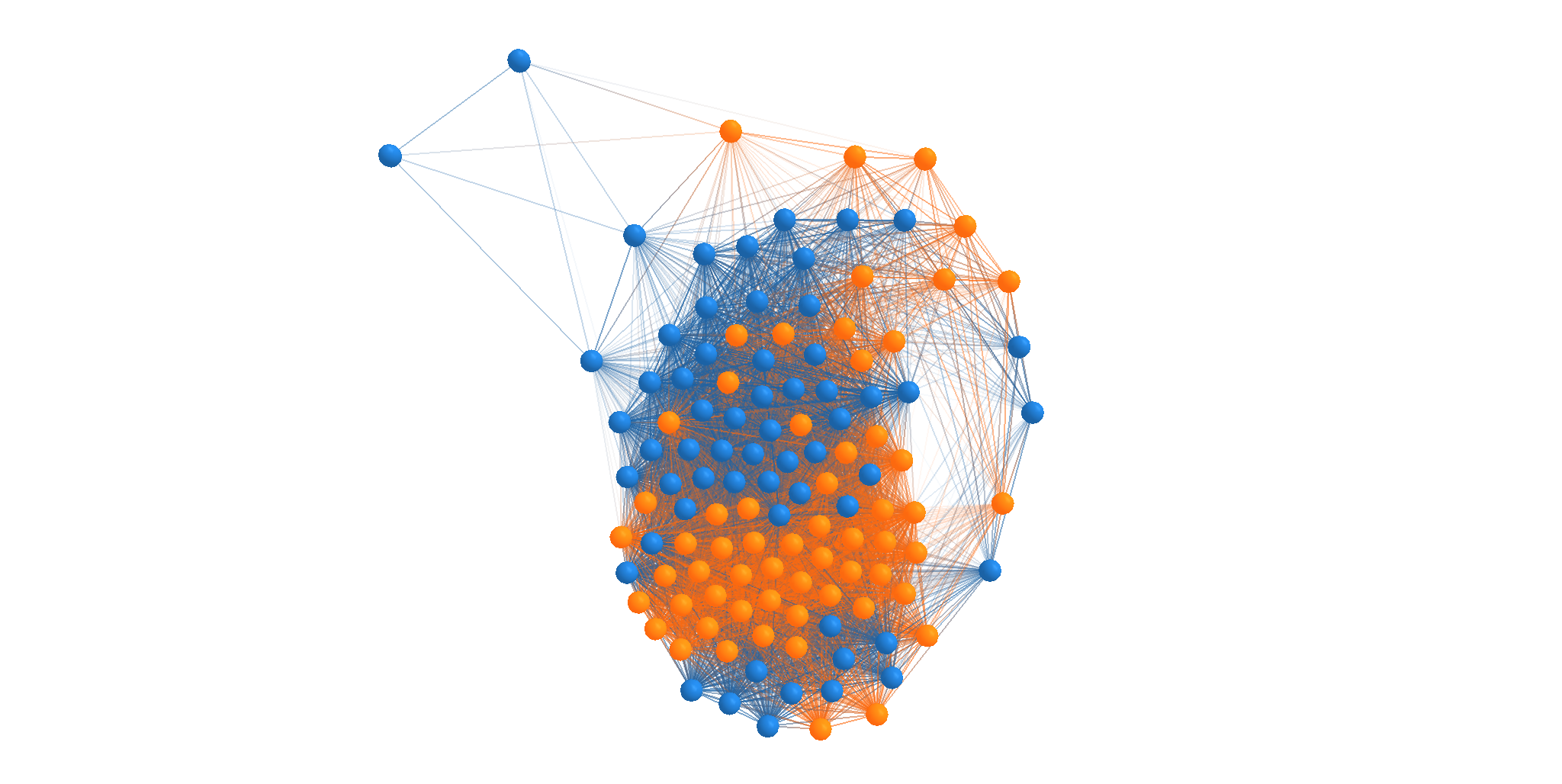}}
    \subfigure[]{\includegraphics[width=0.45\textwidth]{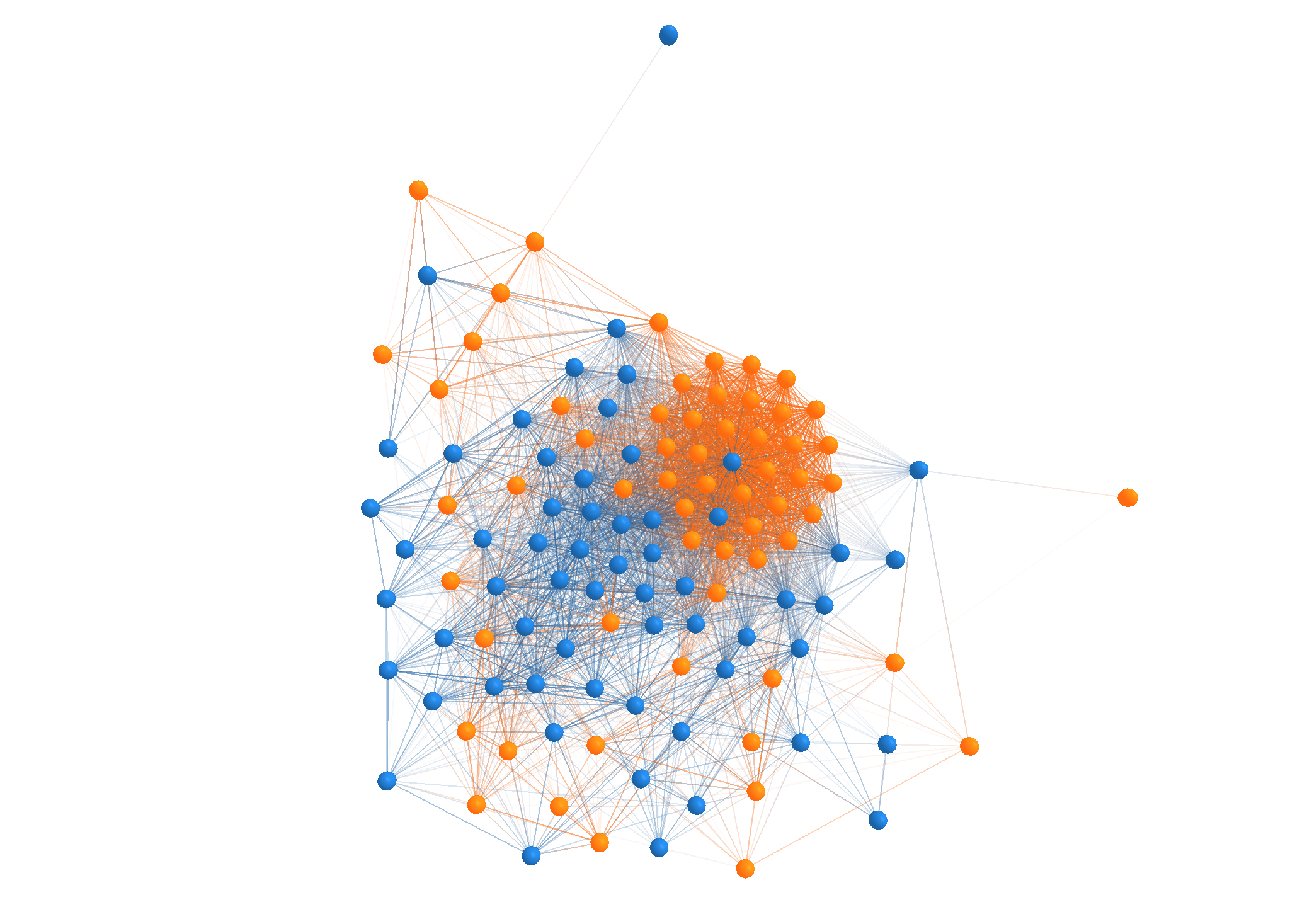}}
    \subfigure[]{\includegraphics[width=0.45\textwidth]{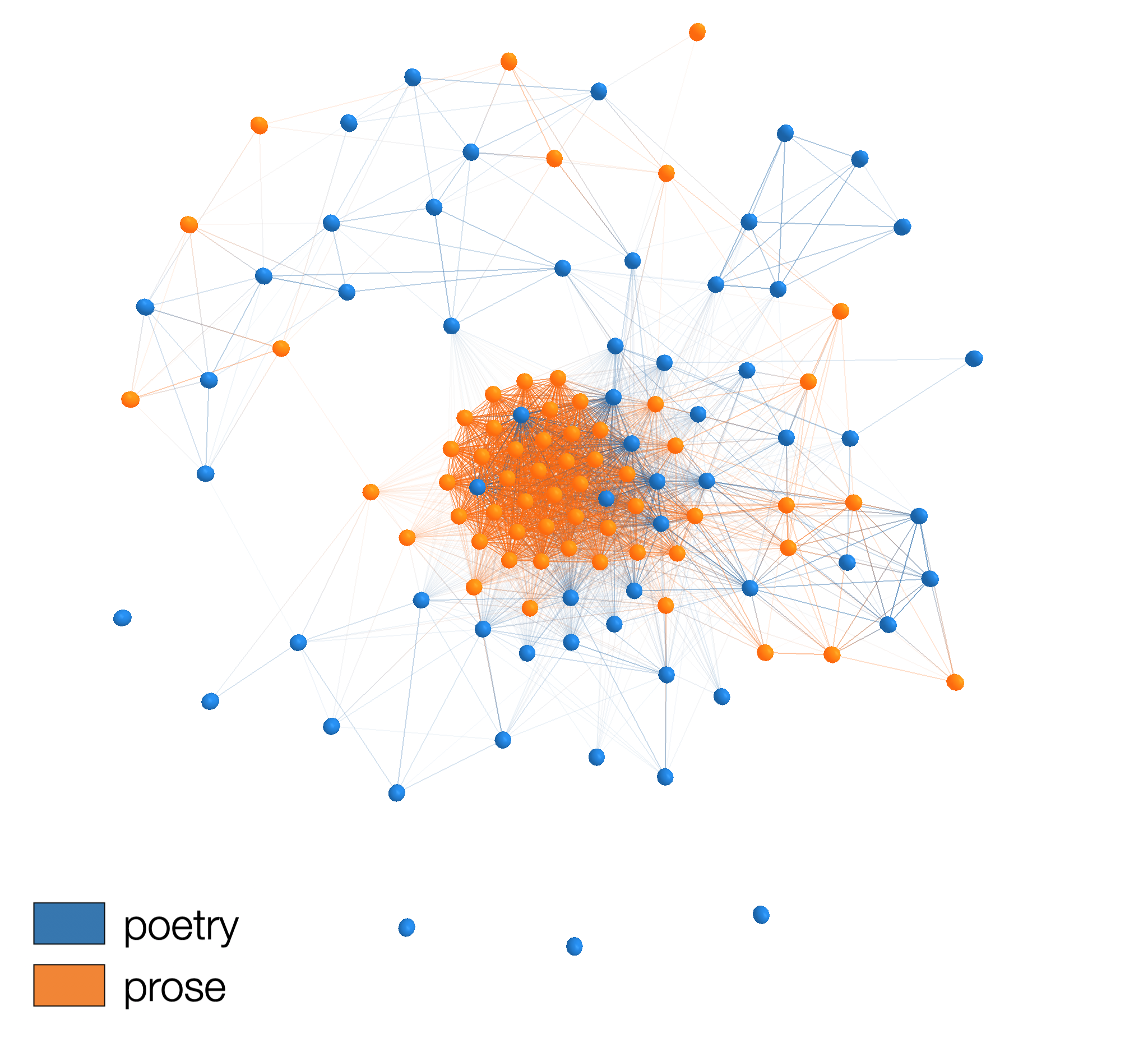}}
    
  \caption{Visualization of the networks, where nodes represent the similarities between the features of each text items (a), (b), and (c) represent 3, 14, and all features, respectively. Here, we removed all edges with weight lower than $\tau =0.5$.}
  \label{fig:networks}
\end{figure}

In Fig.~\ref{fig:networks}, we present the data analysis through a network science approach (as described in Section~\ref{sec:network}). By considering the most relevant features (measured via \emph{information gain}), we depict three different numbers of features. In the first, Fig.~\ref{fig:networks}(a), only the three features used for the LDA classifier were considered. As can be seen, there is natural segregation among nodes of poetry and prose. We also considered 14 features, which gave rise to the best result using the MLP classifier (see Fig.~\ref{fig:networks}(b)). In comparison with the previous case, with 14 features, poetry tends to be much more spread on the visualization. A similar result was found when all features were considered, as shown in Fig.~\ref{fig:networks}(c).
Interestingly, for the cases of Fig.~\ref{fig:networks}(b)~and~(c), poetry was found to be more diverse than prose in terms of its feature vectors. 

\subsection{Comparison with random texts} \label{sec:result_random_texts}

\begin{figure}[b!]
  \centering
    \includegraphics[width=0.49\textwidth]{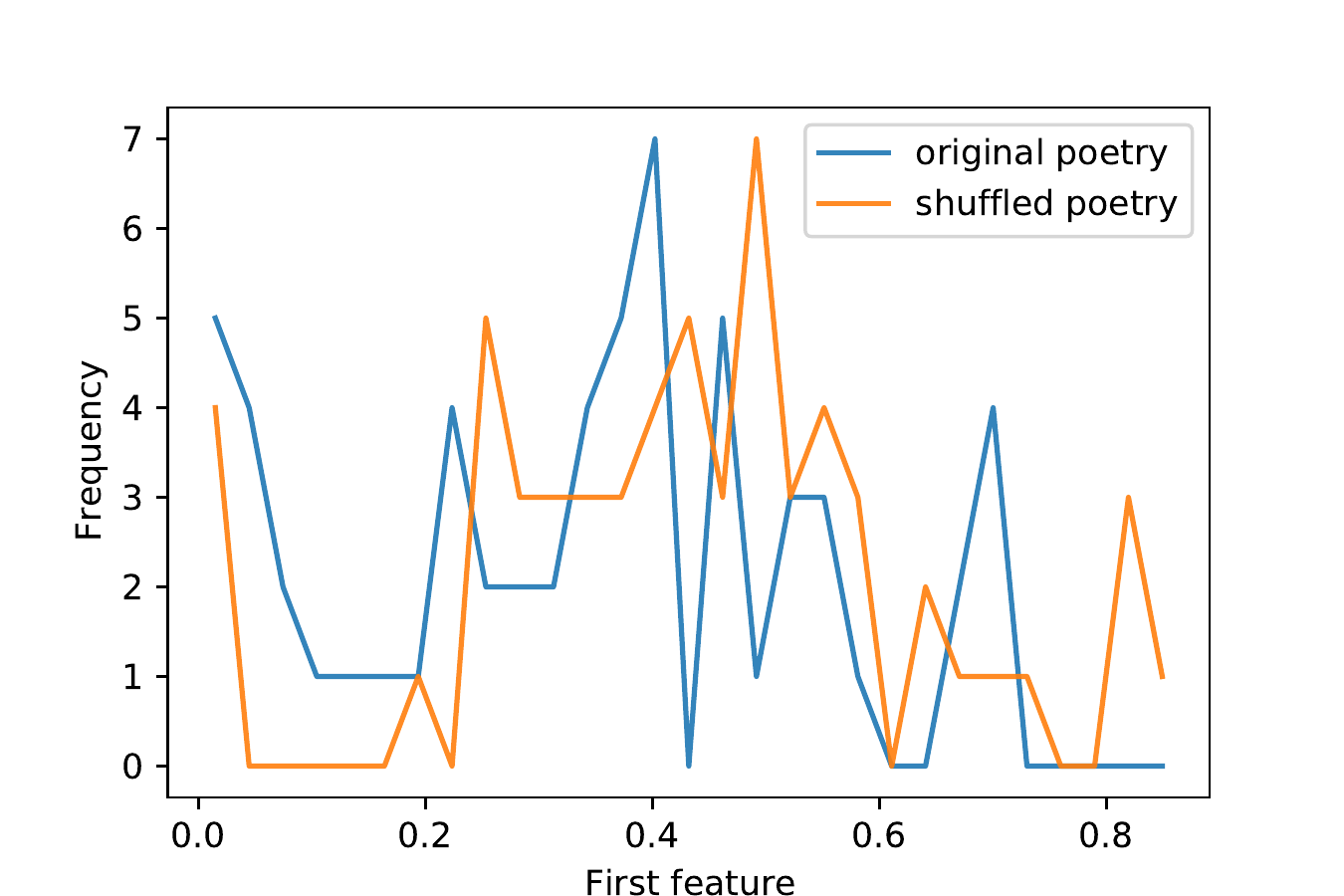}
  \caption{Histogram of the average of \emph{cv}$(l)$, which was considered by \emph{information gain} as being the most important. The employed parameters are: $L_0=2$ and $\Delta=0.15$.}
  \label{fig:feature_poetry_rand}
\end{figure}

In this section, we compare the prose and poetry of original texts with their respectively shuffled versions. We also compared between both shuffled versions. In order to create the shuffled version, we identified all tokens in the text. Next, the position of the punctuation was fixed, and the remaining tokens were shuffled. 

First, we analyze the distribution of the features. Fig.~\ref{fig:feature_poetry_rand} illustrates the first feature selected by \emph{information gain}. Both the samples of original and shuffled poetry are highly overlapping. The next $10$ selected features present similar distributions. This characteristic of shuffled poetry to be similar to the original texts is also reflected on the classification results (see Table~\ref{tab:original_shuffled_poetry}). All in all, the classification quality measurements are much worse than the cases presented in the previous section. 
Furthermore, for all classifiers, a significantly higher number of features was used.
Since in poetry, the author carefully chooses words to create rhymes, we believe that this low accuracy can be result of rhymes randomly generated in the shuffled corpus.

% ------------------------------------------------------------- %
% -------- Comparacao entre poetry and shuffled poetry -------- %
% --------------------------------------------------------------%
\begin{table}[t!]
%\resizebox{0.45\textwidth}{!}
{%
\begin{tabular}{ccccccc}
\hline
                             &                        & \multicolumn{2}{c}{Precision}                                                 & \multicolumn{2}{c}{Recall}                                                    &                            \\
\multirow{-2}{*}{Classifier} & \multirow{-2}{*}{$n_f$} & original                              & shuffled                              & original                              & shuffled                              & \multirow{-2}{*}{Accuracy} \\ \hline
LDA                          & 28                     & \cellcolor[HTML]{DAE8FC}0.62          & \cellcolor[HTML]{FFCE93}\textbf{0.64} & \cellcolor[HTML]{DAE8FC}\textbf{0.67} & \cellcolor[HTML]{FFCE93}0.58          & 0.62                       \\
RF                           & 28                     & \cellcolor[HTML]{DAE8FC}0.63          & \cellcolor[HTML]{FFCE93}0.64          & \cellcolor[HTML]{DAE8FC}0.65          & \cellcolor[HTML]{FFCE93}0.62          & \textbf{0.63}              \\
KNN                          & 26                     & \cellcolor[HTML]{DAE8FC}\textbf{0.64} & \cellcolor[HTML]{FFCE93}0.61          & \cellcolor[HTML]{DAE8FC}0.57          & \cellcolor[HTML]{FFCE93}\textbf{0.68} & 0.62                       \\
SVM                          & 31            & \cellcolor[HTML]{DAE8FC}0.59          & \cellcolor[HTML]{FFCE93}0.61          & \cellcolor[HTML]{DAE8FC}0.63          & \cellcolor[HTML]{FFCE93}0.57          & 0.60                       \\
MLP                          & \textbf{24}                     & \cellcolor[HTML]{DAE8FC}0.61          & \cellcolor[HTML]{FFCE93}0.62          & \cellcolor[HTML]{DAE8FC}0.65          & \cellcolor[HTML]{FFCE93}0.58          & 0.62                       \\ \hline
\end{tabular}%
}
\caption{Performance of classifiers LDA, RF, KNN, SVM and MLP on the original and shuffled poetries.}
\label{tab:original_shuffled_poetry}
\end{table}

% --------------------------------------------------------------%
% --------------------------------------------------------------%
% --------------------------------------------------------------%

In the majority of the selected features, there is a frequency peak close to zero that decreases in the shuffled version, meaning that the shuffled texts may have more rhymes than the original ones.
The difference in the distributions, mainly due to the highest peak, promotes better classification performance. As a result, the overall discriminability metrics are higher than the comparison between original and shuffled poetry (see Fig.~\ref{fig:feature_poetry_rand} and Table~\ref{tab:original_shuffled_poetry}). It is worth noting that the best classifier, RF, obtained an accuracy of $0.72$ with $n_f = 11$. In general, the classifiers needed a lower number of features than previous cases to get meaningful results.

% ------------------------------------------------------------- %
% --------- Comparacao entre prose and shuffled prose --------- %
% --------------------------------------------------------------%

\begin{table}[ht]
%\resizebox{0.45\textwidth}{!}
{%
\begin{tabular}{ccccccc}
\hline
                             &                        & \multicolumn{2}{c}{Precision}                                                 & \multicolumn{2}{c}{Recall}                                                    &                            \\
\multirow{-2}{*}{Classifier} & \multirow{-2}{*}{$n_f$} & original                              & shuffled                              & original                              & shuffled                              & \multirow{-2}{*}{Accuracy} \\ \hline
LDA                          & 7                      & \cellcolor[HTML]{DAE8FC}0.67          & \cellcolor[HTML]{FFCE93}0.75          & \cellcolor[HTML]{DAE8FC}0.80          & \cellcolor[HTML]{FFCE93}0.60          & 0.70                       \\
RF                           & 11            & \cellcolor[HTML]{DAE8FC}\textbf{0.69} & \cellcolor[HTML]{FFCE93}0.76          & \cellcolor[HTML]{DAE8FC}0.80          & \cellcolor[HTML]{FFCE93}\textbf{0.63} & \textbf{0.72}              \\
KNN                          & \textbf{3}              & \cellcolor[HTML]{DAE8FC}0.64          & \cellcolor[HTML]{FFCE93}0.71          & \cellcolor[HTML]{DAE8FC}0.77          & \cellcolor[HTML]{FFCE93}0.57          & 0.67                       \\
SVM                          & 7                      & \cellcolor[HTML]{DAE8FC}0.67          & \cellcolor[HTML]{FFCE93}\textbf{0.77} & \cellcolor[HTML]{DAE8FC}\textbf{0.82} & \cellcolor[HTML]{FFCE93}0.60          & 0.71                       \\
MLP                          & 4                      & \cellcolor[HTML]{DAE8FC}0.67          & \cellcolor[HTML]{FFCE93}0.74          & \cellcolor[HTML]{DAE8FC}0.78          & \cellcolor[HTML]{FFCE93}0.62          & 0.70                       \\ \hline
\end{tabular}%
}
\caption{Performance of classifiers in discriminating original and shuffled prose.}
\label{tab:original_shuffled_prose}
\end{table}

% --------------------------------------------------------------%
% --------------------------------------------------------------%
% --------------------------------------------------------------%

To further investigate the role of the punctuation structure and the words chosen to compose the text, we also compared between shuffled versions of poetry and prose.
It is interesting to see in Table~\ref{tab:shuffled_poetry_prose} that shuffled poetry and proses, which have only their words shuffled but keep their punctuation in the same places as in the original versions, are not that well classified as the original versions (see Table~\ref{tab:poetry_prose}).
The best performance is achieved with LDA, with $n_f = 3$ and accuracy of $0.70$.
This result emphasizes that both text structure and word choice are essential to convey the rhythm that characterizes poetry.
It is worth mentioning that, in this case, the classifier performance is more dependent on the employed classifier. Thus, the features could not discriminate the classes with the same quality as in the comparison between the original texts.

% ------------------------------------------------------------- %
% ---- Comparacao entre shuffled poetry and shuffled prose ---- %
% --------------------------------------------------------------%

\begin{table}[t]
%\resizebox{0.65\textwidth}{!}
{%
\begin{tabular}{ccccccc}
\hline
                             &                        & \multicolumn{2}{c}{Precision}                                                 & \multicolumn{2}{c}{Recall}                                                    &                            \\
\multirow{-2}{*}{Classifier} & \multirow{-2}{*}{$n_f$} & poetry*                               & prose*                                & poetry*                               & prose*                                & \multirow{-2}{*}{Accuracy} \\ \hline
LDA                          & \textbf{3}             & \cellcolor[HTML]{DAE8FC}0.68          & \cellcolor[HTML]{FFCE93}0.57          & \cellcolor[HTML]{DAE8FC}0.38          & \cellcolor[HTML]{FFCE93}0.82          & \textbf{0.70}              \\
RF                           & \textbf{3}             & \cellcolor[HTML]{DAE8FC}\textbf{0.78} & \cellcolor[HTML]{FFCE93}0.60          & \cellcolor[HTML]{DAE8FC}0.42          & \cellcolor[HTML]{FFCE93}\textbf{0.88} & 0.65                       \\
KNN                          & 16                     & \cellcolor[HTML]{DAE8FC}0.65          & \cellcolor[HTML]{FFCE93}\textbf{0.63} & \cellcolor[HTML]{DAE8FC}\textbf{0.60} & \cellcolor[HTML]{FFCE93}0.68          & 0.64                       \\
SVM                          & 17          & \cellcolor[HTML]{DAE8FC}0.68          & \cellcolor[HTML]{FFCE93}0.59          & \cellcolor[HTML]{DAE8FC}0.45          & \cellcolor[HTML]{FFCE93}0.78          & 0.62                       \\
MLP                          & 15                     & \cellcolor[HTML]{DAE8FC}0.67          & \cellcolor[HTML]{FFCE93}0.62          & \cellcolor[HTML]{DAE8FC}0.57          & \cellcolor[HTML]{FFCE93}0.72          & 0.64                       \\ \hline
\end{tabular}%
}
\caption{Performance of classifiers LDA, RF, KNN, SVM and MLP on the shuffled poetry (poetry*) and shuffled proses (prose*).}
\label{tab:shuffled_poetry_prose}
\end{table}

% --------------------------------------------------------------%
% --------------------------------------------------------------%
% --------------------------------------------------------------%

\section{\label{sec:conc} Conclusions}
One of the several features shared by arts and science is their division into major areas or types of works. While in science one may categorize works into physical and biological sciences, a major division in literature concerns the concepts of prose and poetry. While these two important types of works can often be readily identified by humans, the automated classification of respective literary works constitutes a more substantial challenge.  Though rhythm and rhymes are known to be elements typically found in poetry, they also appear to varying degrees in several works understood as prose.  The present work aimed at developing a systematic approach to identifying --- through concepts from network science, pattern recognition and feature selection --- the characteristics that are particularly specific to poetry and prose.  

In the present work, we resorted to prose and poetry texts from the Gutenberg database. We represented the texts in terms of all the identified rhymes and the phones. These representations were characterized in terms of some proposed metrics, including the mean and coefficient of variation of the time intervals, which were then selected through \emph{information gain} attribute selector. In order to test the potential of the features, we employed five different classifiers based on different assumptions. To analyze the results, we also considered a network science-based methodology.

As we developed our methodology, many interesting results were found. First, in the analysis of some basic statistics of the texts (e.g., text size and number of symbols), prose, and poetry were found to be similar. However, by considering the number of rhyme repetitions and the average rhyme repetitions, poetry tends to give rise to a larger diversity of rhymes and repetitions. In the following, by considering the features obtained from the proposed representation and the attribute selection method, the best accuracy result was found for the MLP classifier. 

In order to better understand the relationship between the classes and the features, we represent the relationship between the samples as a complex network. More specifically, the network nodes and links relate to the texts and their feature similarity, respectively. By varying the number of considered features, it was possible to note that poetry rhyme patterns tended to be substantially more diversified than in prose. Even assuming that there is a fixed metric for many of the considered poetry, the result illustrates how diversely poetry can be written. 

Interestingly, the comparison between poetry and shuffled poetry, prose and shuffled prose revealed that the task of classifying between poetry and shuffled poetry is not trivial, which corroborates with the results obtained from the complex network analysis. In other words, the classification task is more challenging since there is a wide range of possibilities for poetry. 

Many possible future works can be developed from the proposed representation and measurements. For instance, one can consider the analysis and comparison between texts of characteristics of literary movements. These features can also be used in more elaborated classification texts combined with other attributes (e.g., word counts).

\section*{Acknowledgments}
H. F. de Arruda acknowledges FAPESP for sponsorship (grants 2018/10489-0). 
S. M. Reia was supported by the Coordena\c{c}\~ao de Aperfei\c{c}oamento de Pessoal de
N\'{\i}vel Superior - Brasil (CAPES) - Finance Code 001.
D. R. Amancio thanks CNPq (grant no.  304026/2018-2). %. and FAPESP (grant no. 20/06271-0). 
L. da F. Costa thanks CNPq (grant no.  307085/2018-0). This work has been supported also by the FAPESP grant 15/22308-2.
H. F. de Arruda thanks Soremartec S.A. and Soremartec Italia, Ferrero Group, for partial financial support (from 1st July 2021). His funders had no role in study design, data collection, and analysis, decision to publish, or manuscript preparation.

\newpage

\bibliographystyle{ieeetr}
\bibliographystyle{abbrv}
\bibliography{ref}






% \end{thebibliography}

\end{document}